\documentclass[10pt,twocolumn,letterpaper]{article}

\usepackage[pagenumbers]{cvpr}
\usepackage{multirow}
\definecolor{cvprblue}{rgb}{0.21,0.49,0.74}
\usepackage[breaklinks,colorlinks,allcolors=cvprblue]{hyperref}

\def\confName{CVPR}
\def\confYear{2027}

\title{Dynamics-Informed Action Learning for Goal-Conditioned Latent Planning}

\author{Yinuo Jiang$^{1}$ \quad Guangzhi Wang\footnotemark[2]\hspace{0.05in} $^{2}$ \quad Yiran Wang\footnotemark[1]\hspace{0.05in} $^{2}$ \quad Junfu Pu$^{2}$ \\
        Cheng Cheng$^{2}$ \quad Honghao Chen$^{2}$ \quad Cheng Cheng\footnotemark[1]\hspace{0.05in} $^{1}$ \quad Ying Shan$^{2}$ \\
        $^{1}$Huazhong University of Science and Technology \\
        $^{2}$Tencent ARC Lab \\
        {\tt\small \{jiangyinuo,c\_cheng\}@hust.edu.cn}
\vspace{-2mm}
}

\begin{document}
\maketitle
\footnotetext[2]{Project Lead.}
\footnotetext[1]{Corresponding Author.}
\begin{abstract}
Latent world models enable visual control by planning action sequences in compact representation spaces, but a gap remains between how latent dynamics are learned and how they are used for planning. First, teacher-forced training does not expose the dynamics model to the recursively predicted latent states encountered during planning. Second, sampling-based planners typically initialize action search from goal-agnostic proposals, without exploiting the action-conditioned dynamics learned by the world model. We introduce Dynamics-Informed Action Learning (DIAL) to bridge these two gaps. During training, rollout alignment exposes the forward model to its own autoregressive latent predictions, while multi-step inverse supervision trains a lightweight inverse dynamics head to infer actions from latent-state transitions. We further apply the inverse head to model-generated transitions with its parameters held fixed, encouraging predicted rollouts to retain action-relevant information. At test time, DIAL autoregressively couples the inverse and forward dynamics to construct a dynamics-informed action sequence toward the goal, which initializes the mean of the Cross-Entropy Method (CEM) proposal distribution. The subsequent CEM search then refines this initialization using the standard sampling and updating procedure. Experiments across four visual control benchmarks demonstrate that DIAL outperforms existing latent world models and planning baselines in both standard and long-horizon scenarios.
\end{abstract}    
\section{Introduction}
\label{sec:intro}

World models enable agents to predict the consequences of their actions before execution~\cite{ha2018recurrent,hafner2019learning,hafner2019dream}, allowing them to learn from imagined rollouts rather than environment interactions. This paradigm has demonstrated strong performance across diverse control settings, with recent world-model agents achieving broad task coverage and performance surpassing human levels~\cite{hafner2023mastering,hafner2025training}. In visual planning, latent world models bypass pixel-level reconstruction and operate directly in a compact latent space~\cite{maes2026leworldmodel}. During training, forward dynamics models learn how actions transform the current latent state into a future state. At planning time, goal-conditioned control generates candidate action sequences and selects those whose predicted outcomes best match the goal~\cite{sun2026intact}. However, this forward model
defines a one-way, action-conditioned mapping from actions to state transitions: it predicts the resulting state given an action, but does not directly infer which action would induce a desired latent-state transition. This asymmetry creates a gap between learned dynamics and planning.

\begin{figure*}[t]
\centering
\includegraphics[width=\linewidth]{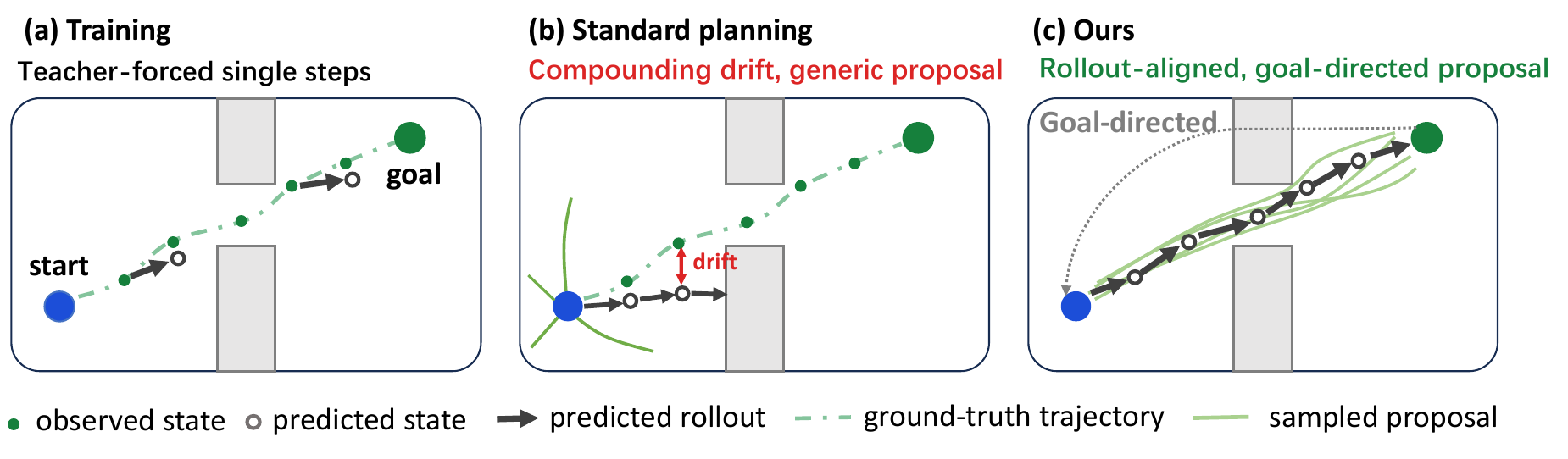}
\vspace{-8mm}
\caption{\textbf{DIAL aligns latent dynamics learning with goal-directed planning.}
(a) During training, teacher forcing performs one-step prediction from observed latent states.
(b) At planning time, autoregressive rollouts can accumulate prediction drift, while generic action proposals lack goal-directed initialization.
(c) DIAL addresses both gaps by training on model-generated rollouts and using learned inverse dynamics to construct a dynamics-informed CEM proposal.}
\label{fig:gap}
\vspace{-4mm}
\end{figure*}

This asymmetry appears in two complementary ways. First, there is an inconsistency in the state inputs received by the model during the training and planning phases. During training, each prediction relies on latent states derived from encoded observation frames. Conversely, during planning, the model must autoregressively feed its own predictions into subsequent steps, causing errors to accumulate during the rollout process~\cite{li2026predictive}. Consequently, accurate single-step prediction does not necessarily translate into reliable multi-step planning. Second, existing planning methods fail to fully leverage the action-conditioned dynamics learned by the world model when initializing the action search. Planners typically initialize candidate action sequences based on generic uniform or Gaussian distributions~\cite{de2005tutorial,sun2026intact}, without using this learned information to identify promising regions of the action space toward the goal. As a result, under limited search budgets or large goal offsets, such uninformed initialization forces early iterations to explore irrelevant regions of the action space, reducing search efficiency.

Existing approaches address these challenges mainly from the modeling and trajectory-evaluation perspectives. On the modeling side, methods such as DINO-WM~\cite{zhou2024dino}, PLDM~\cite{sobal2022joint,sobal2025stress}, and LeWM~\cite{maes2026leworldmodel} improve latent representations and dynamics modeling. On the trajectory-evaluation side, methods such as RC-aux~\cite{li2026predictive} and TRM~\cite{li2026beyond} improve candidate evaluation. These approaches, however, do not directly connect learned action-conditioned dynamics to the initialization of goal-conditioned action search. As illustrated in Fig.~\ref{fig:gap}, autoregressive prediction and action-search initialization therefore remain two separate aspects of the planning pipeline.

To this end, we propose \textbf{D}ynamics-\textbf{I}nformed \textbf{A}ction \textbf{L}earning (\textbf{DIAL}) for latent world models. DIAL connects latent dynamics learning with autoregressive planning and uses learned inverse dynamics to inform the initialization of action search. During training, rollout alignment exposes the forward model to its own predicted latent states, while multi-step inverse supervision trains a lightweight inverse dynamics head to recover actions from latent-state transitions. We further apply the inverse head to model-generated transitions with the forward model held fixed, encouraging rollouts to preserve action-relevant information. During planning, the inverse head infers an action toward the goal from the current predicted latent state, and the forward model predicts the resulting latent state for the next step. The resulting action sequence initializes only the mean of the Cross-Entropy Method (CEM) proposal distribution, while the subsequent sampling and proposal-refitting procedure remains unchanged. DIAL therefore transfers action-conditioned information learned during world-model training into goal-directed search without introducing an additional planning stage or increasing the search budget.

We instantiate DIAL on LeWorldModel (LeWM), a compact reconstruction-free JEPA world model for goal-conditioned planning from pixels~\cite{maes2026leworldmodel}. DIAL preserves the original LeWM backbone and CEM planning procedure, allowing us to isolate the effects of rollout-aligned training and dynamics-informed action initialization. We evaluate DIAL on the original LeWM task family under matched training and planning protocols, including different planning horizons, search budgets, and ablation settings. DIAL consistently improves goal-reaching performance, with particularly pronounced gains under limited search budgets and long-horizon goals. These results demonstrate the benefit of combining rollout-aligned latent dynamics with dynamics-informed action initialization for efficient latent planning. Our contributions can be summarized as:

\begin{enumerate}[leftmargin=*,itemsep=2pt]

\item We propose DIAL, a dynamics-informed framework that connects latent dynamics learning with goal-conditioned planning by using learned inverse dynamics to construct initial CEM proposal.

\item We introduce rollout alignment and multi-step inverse supervision to improve autoregressive dynamics and preserve action-relevant information in model-generated latent transitions.

\item We provide controlled empirical evaluation across multiple goal offsets, search budgets, and training ablations, demonstrating consistent improvements in goal-reaching performance, particularly under limited search budgets and long-horizon planning.

\end{enumerate}

\section{Related Work}
\label{sec:formatting}

\paragraph{World models for representation.}
World models aim to learn predictive models of environment dynamics from data, enabling agents to reason about future states through imagined rollouts~\cite{ouyang2026player}. Early approaches rely on generative models that explicitly model future observations, such as IRIS~\cite{micheli2022transformers}, DIAMOND~\cite{alonso2024diffusion}, and DreamerV4~\cite{hafner2025training}. More recently, Joint-Embedding Predictive Architectures (JEPAs) have emerged as a reconstruction-free alternative, learning predictive dynamics directly in a compact latent space without explicit pixel reconstruction. Within action-conditioned latent world modeling, JEPA-based methods differ in how representations are learned. Methods based on frozen pretrained encoders, such as DINO-WM~\cite{zhou2024dino}, constrain the learned state space through pre-trained visual representations~\cite{wang2025tacodepth,jiang2026spectral}. End-to-end approaches such as PLDM~\cite{sobal2022joint,sobal2025stress}, while capable of jointly learning representations and dynamic information from raw observation data~\cite{wang2023neural,wang2024nvds}, may face training instability issues. LeWM~\cite{maes2026leworldmodel} addresses this challenge with a stable end-to-end framework that combines latent prediction with representation regularization. Our method is based on the end-to-end JEPA framework and investigates how to leverage the learned dynamic model to improve planning.

\paragraph{Candidate evaluation.}
Beyond improving learned representations and dynamics models, another line of work improves planning by refining the objective used to evaluate candidate trajectories. Fast-LeWM~\cite{gao2026fast} introduces a self-consistency score during candidate selection to filter out dynamical hallucinations, while RC-aux~\cite{li2026predictive} and TRM~\cite{li2026beyond} augment the terminal cost with reachability-aware metrics or temporal distance constraints. These approaches primarily improve the evaluation and ranking of generated candidates, while leaving the  proposal distribution unchanged. Consequently, when the search budget is limited or the task requires substantial exploration~\cite{wang2026gamehorizon}, the quality of generated candidate set may become a bottleneck that cannot be resolved solely through improved candidate evaluation.

\paragraph{Inverse dynamics and action priors.}

Learned inverse dynamics and action priors have also been explored for visual planning~\cite{wang2026taskframe}. GC-IDM~\cite{nguyen2026latent}, for example, uses inverse dynamics to directly generate goal-conditioned actions, making planning dependent on the accuracy of the learned action predictor. PRISM~\cite{wang2026prism} learns a goal-conditioned Gaussian prior separately from a frozen world model, leaving the action prior decoupled from the learned dynamics. INTACT~\cite{sun2026intact} incorporates learned intent-to-action mappings into CEM-based~\cite{rubinstein2004cross} planning, primarily using the learned mapping for direct action generation or local search verification. In contrast, DIAL jointly learns inverse dynamics with the latent world model and uses the resulting dynamics-informed action sequence to initialize the CEM proposal mean, while retaining CEM for subsequent candidate sampling and refinement.

\section{Preliminaries}
\label{sec:prelim}

\paragraph{Latent world model.}
Following LeWM~\cite{maes2026leworldmodel}, we consider a reconstruction-free
latent world model consisting of a visual encoder
$E_\phi:\mathcal{O}\rightarrow\mathbb{R}^D$ and an action-conditioned latent
dynamics model $F_\theta$. The encoder maps observations to latent states:
\begin{equation}
z_t = E_\phi(o_t).
\end{equation}
The dynamics model predicts the next latent state from a temporal context of latent states and actions:
\begin{equation}
\hat z_{t+1}
=
F_\theta
\!\left(
z_{t-h+1:t},
a_{t-h+1:t}
\right),
\end{equation}
where $h$ denotes the context length and $a_t$ is the current action used to predict $\hat z_{t+1}$. Following the reconstruction-free formulation of LeWM, the model is trained to predict future latent states from observed trajectories together with an anti-collapse regularizer. At planning time, the learned dynamics model is applied autoregressively to predict the consequences of candidate action sequences.

\paragraph{Goal-conditioned planning.}
We consider finite-horizon goal-conditioned control from visual observations. At time $t$, the agent observes $o_t$ and is given a goal observation $o_g$, with corresponding latent states $z_t$ and $z_g$. Given the current latent context and goal latent, planning seeks an action sequence whose predicted terminal latent is closest to the goal:
\begin{equation}
\begin{aligned}
a_{t:t+H-1}^{*}
\in
\arg\min_{a_{t:t+H-1}}
d\!\left(
\tilde z_{t+H},
z_g
\right),
\\
\tilde z_{t+H}
=
F_\theta^{(H)}
\!\left(
z_{t-h+1:t},
a_{t-h+1:t+H-1}
\right),
\end{aligned}
\end{equation}
where $d(\cdot,\cdot)$ denotes the latent goal-matching cost and
$F_\theta^{(H)}$ denotes the autoregressive $H$-step rollout of
$F_\theta$. The historical actions $a_{t-h+1:t-1}$ are fixed context, while the future actions $a_{t:t+H-1}$ are optimization variables. Planning is typically implemented using sampling-based Model Predictive Control methods, such as the Cross-Entropy Method (CEM)~\cite{rubinstein2004cross}. CEM samples candidate action sequences from a Gaussian proposal distribution, evaluates their terminal latent costs, selects the top $N_{\mathrm{e}}$ elite sequences, and refits the proposal distribution using these elites. This process is repeated for a fixed number of iterations, after which the resulting action sequence is executed in a receding-horizon manner.

\section{Method}
\label{sec:method}

\begin{figure*}[t]
    \centering
    \includegraphics[width=\linewidth]{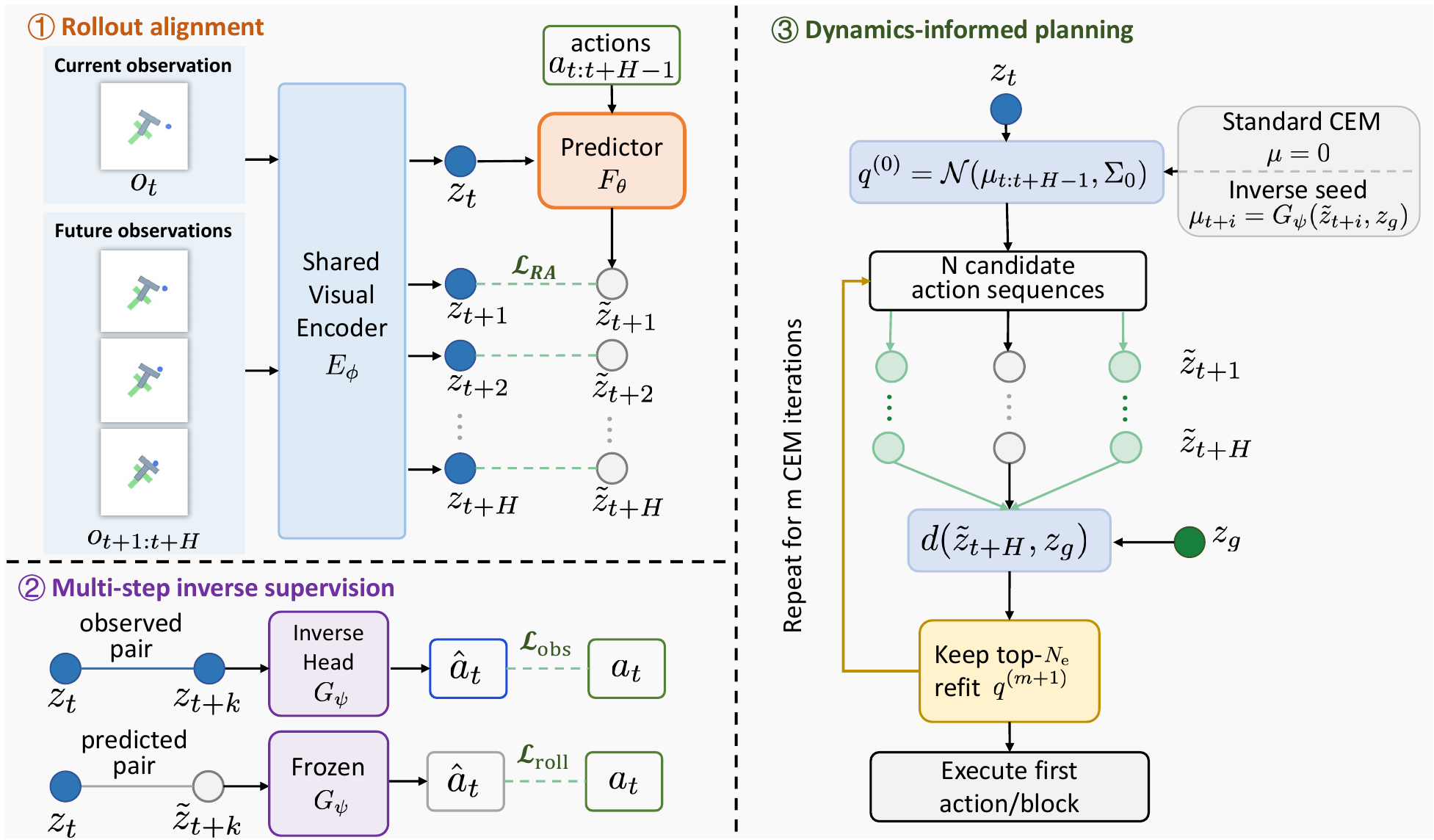}
    \vspace{-4mm}
\caption{Overview of DIAL. \textbf{(1) Rollout alignment} trains the latent dynamics predictor $F_\theta$ on autoregressively generated latent contexts to reduce the mismatch between training and planning. \textbf{(2) Multi-step inverse supervision} trains the inverse-dynamics head $G_\psi$ on observed latent state pairs and uses a frozen inverse head $G_{\bar\psi}$ to encourage model-generated states to preserve action-relevant information. \textbf{(3) Dynamics-informed planning} autoregressively couples $G_\psi$ and $F_\theta$ to generate a goal-directed action sequence, which initializes only the mean of the CEM proposal.}
    \vspace{-4mm}
    \label{fig:overview}
\end{figure*}

\subsection{Overview}
\label{sec:method-overview}

We propose Dynamics-Informed Action Learning (DIAL) for latent world models, a framework that aligns latent dynamics learning with autoregressive planning and leverages learned inverse dynamics to initialize action search. As illustrated in Fig.~\ref{fig:overview}, DIAL comprises a visual encoder $E_\phi$, an action-conditioned latent predictor $F_\theta$, and a lightweight inverse dynamics head $G_\psi$. The encoder maps observations to $D$-dimensional latent states, while the predictor $F_\theta$ forecasts future latent states based on the current latent state and action context. The proposed inverse dynamics head $G_\psi$ is responsible for estimating the action connecting a pair of latent states.

Building on the base world-model objective, DIAL introduces three complementary training terms. First, rollout alignment trains $F_\theta$ on autoregressively generated latent contexts, reducing the mismatch between teacher-forced training and planning-time rollouts. Second, multi-step inverse supervision trains $G_\psi$ to recover actions from observed latent state pairs. Third, inverse supervision on predicted states applies a frozen copy of the inverse head to model-generated latent pairs, encouraging autoregressive rollouts to preserve action-relevant information. At planning time, DIAL uses $G_\psi$ and $F_\theta$ autoregressively to construct a goal-directed action sequence. Starting from the current latent state, $G_\psi$ estimates an action toward the goal from the current latent state and goal latent, and $F_\theta$ advances the latent state under the predicted action. The resulting sequence initializes only the mean of the CEM proposal distribution, while CEM retains its standard sampling, evaluation, elite selection, proposal refitting, and search budget.

\subsection{Rollout-Aligned Latent Training}
\label{sec:method-train}

\paragraph{Rollout alignment.}
The base prediction objective trains $F_\theta$ with teacher forcing, where each prediction is conditioned on latent states encoded from observed frames. During planning, however, $F_\theta$ is applied autoregressively, so its own predictions become inputs to subsequent steps. To align these two input regimes, we train $F_\theta$ over an autoregressive rollout of depth $L$. Starting from the observed latent context $\tilde z_{t-h+1:t}=z_{t-h+1:t}$, we generate an autoregressive rollout by feeding each predicted latent state back into the predictor. Specifically, the $j$-th predicted state is obtained as:
\begin{equation}
\begin{aligned}
\tilde z_{t+j}
=
F_\theta\!\left(
\tilde z_{t+j-h:t+j-1},
a_{t+j-h:t+j-1}
\right),
\\
j=1,\ldots,L,
\label{eq:ra-rollout}
\end{aligned}
\end{equation}

The first prediction is therefore conditioned entirely on observed latents, while subsequent input windows progressively replace observed states with model predictions. For $j>h$, the predictor is conditioned entirely on model-generated latents. We then minimize the rollout alignment loss as:
\begin{equation}
\mathcal{L}_{\mathrm{RA}}
=
\frac{1}{L}
\sum_{j=1}^{L}
\left\|
\tilde z_{t+j}
-
\mathrm{sg}[z_{t+j}]
\right\|_2^2,
\label{eq:ra-loss}
\end{equation}
where $\mathrm{sg}[\cdot]$ denotes stop-gradient. It is applied only to the target latents, while gradients propagate through complete autoregressive recursion. Thus, $\mathcal{L}_{\mathrm{RA}}$ directly trains $F_\theta$ on model-generated inputs encountered during planning.

\paragraph{Multi-step inverse supervision.}
Rollout alignment improves the reliability of autoregressive states, but does
not explicitly require these states to preserve information about the actions
that caused them. We therefore introduce a lightweight inverse dynamics head
$G_\psi$ to recover the initiating action from a latent-state transition. For
an ordered transition from $z_t$ to $z_{t+k}$, we predict
\begin{equation}
\hat a_t = G_\psi(z_t,z_{t+k}),
\qquad
k=1,\ldots,K_{\mathrm{obs}},
\end{equation}
where the supervision target is always the first action $a_t$. Thus, for
$k>1$, the head recovers the action initiating the transition rather than the
complete intervening action sequence. We train $G_\psi$ on latent-state pairs encoded from observed
trajectories:
\begin{equation}
\mathcal{L}_{\mathrm{obs}}
=
\sum_{k=1}^{K_{\mathrm{obs}}}
\omega_k
\left\|
G_\psi(z_t,z_{t+k})-a_t
\right\|_2^2,
\label{eq:inv-observed}
\end{equation}
where $\omega_k\propto K_{\mathrm{obs}}-k+1$ and
$\sum_k\omega_k=1$. This objective trains the inverse mapping while
encouraging the shared latent representation to retain information about the initiating action. 

We further impose the same recoverability constraint on autoregressively
predicted states. Let $\tilde z_{t+k}$ denote a state generated by the
rollout in Eq.~(\ref{eq:ra-rollout}). We reuse $G_\psi$ on these model-generated
pairs while holding its parameters fixed:
\begin{equation}
\mathcal{L}_{\mathrm{roll}}
=
\frac{1}{K_{\mathrm{roll}}}
\sum_{k=1}^{K_{\mathrm{roll}}}
\left\|
G_\psi\!\left(
\mathrm{sg}[z_t],
\tilde z_{t+k}
\right)-a_t
\right\|_2^2.
\label{eq:inv-rollout}
\end{equation}

During this loss, the parameters of $G_\psi$ are held fixed, so gradients flow through its latent inputs and the autoregressive rollout rather than through the inverse mapping itself. The anchor latent is also detached to prevent this loss from directly updating the first input of $G_\psi$. Consequently, $\mathcal{L}_{\mathrm{obs}}$ learns action recovery from observed transitions, while $\mathcal{L}_{\mathrm{roll}}$ encourages model-generated states to preserve the same action-relevant information.

\paragraph{Overall training objective.}
In addition to the objectives above, we retain the reconstruction-free latent prediction objective of LeWM~\cite{maes2026leworldmodel}:
\begin{equation}
\mathcal{L}_{\mathrm{pred}}
=
\left\|
F_\theta\!\left(
z_{t-\ell+1:t},
a_{t-\ell+1:t}
\right)
-
z_{t+1}
\right\|_2^2.
\label{eq:base-pred}
\end{equation}

The predictor accepts variable-length contexts, and this objective is additionally averaged over all valid context lengths $1\leq\ell\leq h$ rather than supervising only the longest one. Following LeWM, we additionally employ the Sketch Isotropic Gaussian Regularizer (SIGReg) to prevent representation collapse:
\begin{equation}
\mathcal{L}_{\mathrm{WM}}
=
\mathcal{L}_{\mathrm{pred}}
+
\lambda_{\mathrm{sig}}\mathcal{L}_{\mathrm{SIG}}.
\label{eq:base-wm}
\end{equation}

Combining this objective with the DIAL-specific terms yields the full training objective:
\begin{equation}
\mathcal{L}_{\mathrm{DIAL}}
=
\mathcal{L}_{\mathrm{WM}}
+
\beta_{\mathrm{RA}}\mathcal{L}_{\mathrm{RA}}
+
\beta_{\mathrm{obs}}\mathcal{L}_{\mathrm{obs}}
+
\beta_{\mathrm{roll}}\mathcal{L}_{\mathrm{roll}},
\label{eq:method-objective}
\end{equation}
where $\beta_{\mathrm{RA}}$, $\beta_{\mathrm{obs}}$, and $\beta_{\mathrm{roll}}$ control the weights of the three DIAL-specific training terms.

\subsection{Dynamics-Informed Planning}
\label{sec:method-inference}

Standard CEM initializes its proposal with a goal-agnostic Gaussian distribution \(q^{(0)}=\mathcal{N}(\mathbf{0},\Sigma_0)\). At planning time, DIAL uses the learned inverse dynamics head to construct a dynamics-informed proposal sequence. Starting from the current latent state \(\tilde z_t = z_t\), the inverse head recursively predicts an action conditioned on the current predicted state and the goal latent, and the forward model advances the latent state under the predicted action:
\begin{equation}
\begin{aligned}
\mu_{t+i}
&=
G_\psi\!\left(\tilde z_{t+i},z_g\right), \\
\tilde z_{t+i+1}
&=
F_\theta\!\left(\tilde z_{t+i},\mu_{t+i}\right),
\qquad i=0,\ldots,H-1.
\end{aligned}
\label{eq:isp-seed}
\end{equation}

For notational simplicity, Eq.~(\ref{eq:isp-seed}) writes \(F_\theta\) with a single state and action, while in practice the predictor retains the most recent \(h\) latent states and actions, following the same convention as during training. Each action is therefore inferred from the latest predicted latent state, producing a goal-conditioned proposal sequence \(\mu_{t:t+H-1}\). Because \(z_g\) may lie beyond the temporal offsets used to train \(G_\psi\), this sequence is not assumed to be an accurate open-loop plan. This sequence is used only to initialize the mean of the CEM proposal:
\begin{equation}
q^{(0)}
=
\mathcal{N}\!\left(\mu_{t:t+H-1},\Sigma_0\right),
\label{eq:isp-proposal}
\end{equation}
where the initial covariance \(\Sigma_0\) remains unchanged. CEM then proceeds with its standard sampling, model rollout, elite selection, and proposal refitting steps. The inverse-decoded sequence only biases the initial search distribution and is never executed directly. Subsequent CEM iterations can therefore correct errors in the initialization.

\section{Experiments}
\label{sec:experiments}

\subsection{Experimental Setup}
\label{sec:exp-setup}

\paragraph{Environments and datasets.}
We follow the evaluation protocol of LeWM~\cite{maes2026leworldmodel} and use the same offline datasets, observation preprocessing, and goal-conditioned planning setup. The environments include TwoRoom~\cite{sobal2025stress}, PushT~\cite{zhou2024dino}, Reacher~\cite{tassa2018deepmind}, and Cube~\cite{park2025ogbench}. Each episode provides an initial observation and a goal image. We report success rate as the primary metric. We use a standard goal offset of 25 steps and additionally introduce an offset-50 setting to evaluate planning over a substantially larger temporal distance. Within each evaluation setting, all methods use the same interaction and evaluation protocol.

\begin{table*}[t]
  \centering
  \small
  \caption{\textbf{Success rates (\%) at the standard 25-step and long-horizon 50-step goal offsets.} Offset-25 results are quoted from the corresponding papers unless otherwise noted. Offset-50 results are evaluated using a unified harness with identical episodes (\(n=50\)) and the same CEM candidate budget across methods. Bold and underline indicate the best and second-best results.}

  \label{tab:combined}
  \resizebox{\textwidth}{!}{%
  \begin{tabular}{l rrrrr rrrrr}
  \toprule
  \multirow{2}{*}{Method} & \multicolumn{5}{c}{Offset 25} & \multicolumn{5}{c}{Offset 50} \\
  \cmidrule(lr){2-6} \cmidrule(lr){7-11}
   & PushT & Cube & Reacher & TwoRoom & Macro & PushT & Cube & Reacher & TwoRoom & Macro \\
  \midrule
  PLDM~\cite{sobal2022joint,sobal2025stress} & 78.0 & 65.0 & 78.0 & 97.0 & 79.5 & 20.0 & 46.0 & 66.0 & 78.0 & 52.5 \\
  DINO-WM~\cite{zhou2024dino} & 74.0 & 86.0 & 79.0 & \textbf{100.0} & 84.8 & 14.0 & 46.0 & 20.0 & \underline{94.0} & 43.5 \\
  LeWM~\cite{maes2026leworldmodel} & \underline{96.0} & 74.0 & 86.0 & 87.0 & 85.8 & 60.0 & 64.0 & 72.0 & 56.0 & 63.0 \\
  RC-aux~\cite{li2026predictive} & 90.8 & 76.0 & 87.2 & \underline{98.0} & 88.0 & \textbf{64.0} & 54.0 & 82.0 & 54.0 & 63.5 \\
  Fast-LeWM~\cite{gao2026fast} & \underline{96.0} & 80.0 & \underline{88.0} & \underline{98.0} & 90.5 & 46.0 & 58.0 & 80.0 & 50.0 & 58.5 \\
  Qantara~\cite{rakhimov2026qantara} & 90.1 & 93.7 & 80.9 & \textbf{100.0} & 91.2 & 40.0 & \underline{68.0} & 78.0 & 84.0 & 67.5 \\
  Fast-LeWM-SC~\cite{gao2026fast} & \textbf{98.0} & 82.0 & \textbf{90.0} & \underline{98.0} & 92.0 & 50.0 & 62.0 & 78.0 & 50.0 & 60.0 \\
  INTACT (Pure CEM)~\cite{sun2026intact} & 88.4 & 68.4 & 83.7 & 82.9 & 80.9 & 54.0 & 46.0 & 72.0 & 38.0 & 52.5 \\
  INTACT (Actor CEM)~\cite{sun2026intact} & 93.6 & \textbf{96.9} & 86.6 & \underline{98.0} & \underline{93.8} & \underline{62.0} & 48.0 & \underline{86.0} & \textbf{96.0} & \underline{73.0} \\
  \midrule
  \textbf{Ours} & \textbf{98.0} & \underline{96.0} & \textbf{90.0} & \textbf{100.0} & \textbf{96.0} & \underline{62.0} & \textbf{70.0} & \textbf{88.0} & 92.0 & \textbf{78.0} \\
  \bottomrule
  \end{tabular}%
  }
\end{table*}

\paragraph{Planning protocol.}

Following prior work~\cite{maes2026leworldmodel,gao2026fast,li2026predictive}, we use an action block size of 5 and a planning horizon of $H=5$, so each plan spans 25 primitive environment steps. An offset of 25 can therefore be addressed within a single plan, whereas an offset of 50 requires replanning while keeping the per-plan horizon unchanged. Importantly, the long-horizon setting does not increase the CEM search budget or planning horizon. The dynamics-informed sequence is generated once to initialize the CEM proposal mean and does not introduce an additional CEM iteration or candidate rollout. Unless otherwise stated, all methods use the same CEM sampling budget, update rule, and replanning schedule within each evaluation setting.

\paragraph{Baselines.} 
We compare DIAL against prior methods that modify different components of the same goal-conditioned latent planning pipeline. Specifically, PLDM~\cite{sobal2022joint,sobal2025stress}, DINO-WM~\cite{zhou2024dino}, LeWM~\cite{maes2026leworldmodel}, Fast-LeWM~\cite{gao2026fast}, and Qantara~\cite{rakhimov2026qantara} modify the learned representations or the forward dynamics. RC-aux~\cite{li2026predictive} modifies the candidate re-scoring mechanism. INTACT~\cite{sun2026intact} replaces the action proposal distribution with learned action mappings or priors. Among these, LeWM is the closest point of comparison because it shares the same backbone architecture and planning protocol. Where official reports are available, baseline results are cited directly from the original papers. For settings not reported in the original publications, we evaluate methods using officially released checkpoints or retrain them using publicly available implementations, adhering strictly to the standardized evaluation protocol for each benchmark.

\begin{figure*}[t]

\centering

\includegraphics[width=\linewidth]{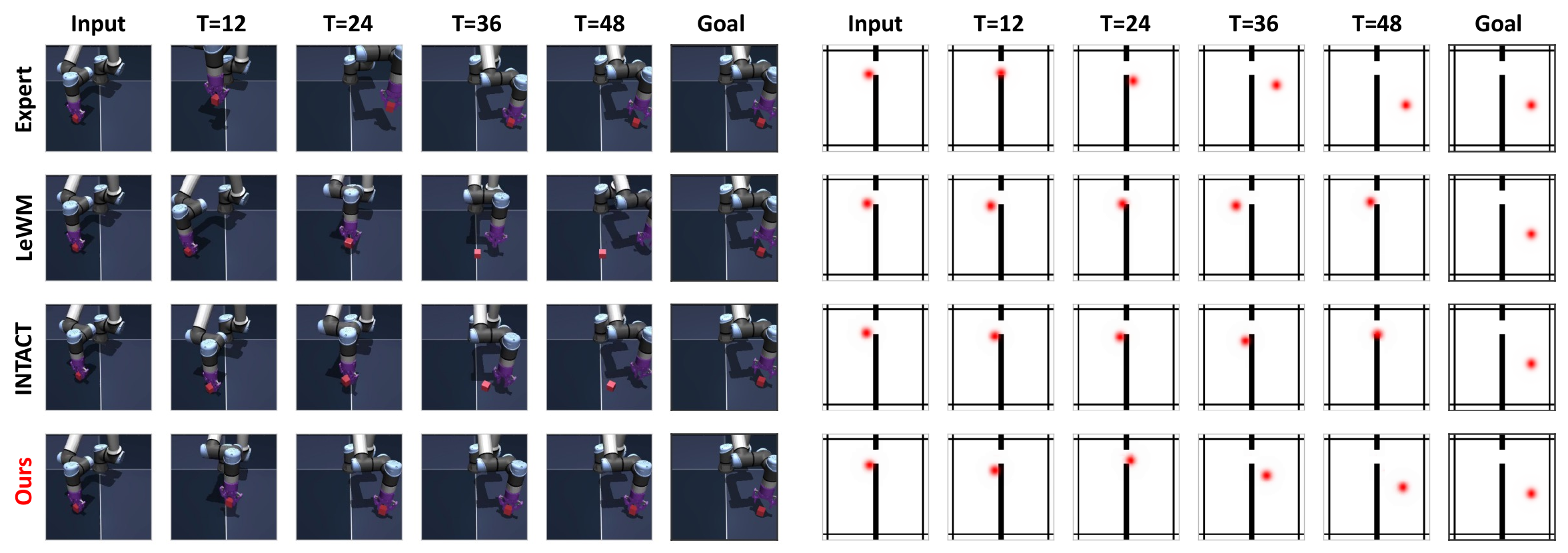}
\vspace{-8mm}
\caption{\textbf{Visualization of planning trajectories on Cube and PushT.} Compared with LeWM and INTACT, DIAL maintains object interaction and makes greater progress toward the goal, while other methods either lose object interaction or terminate far from target.}
\label{fig:qualitative}
\end{figure*}

\vspace{-8mm}
\subsection{Main Results}
\label{sec:exp-main}

\paragraph{Comparison at the standard 25-step offset.} The left part of Table~\ref{tab:combined} summarizes performance at the standard 25-step offset. DIAL achieves the highest macro success rate of \(96.0\%\). Compared with prior-free LeWM~\cite{maes2026leworldmodel}, DIAL improves the macro score by \(10.2\) points, with particularly large gains of \(22.0\) and \(13.0\) points on Cube~\cite{park2025ogbench} and TwoRoom~\cite{sobal2025stress}, respectively. Compared with Fast-LeWM~\cite{gao2026fast} and RC-aux~\cite{li2026predictive}, which improve candidate evaluation while retaining prior-free proposal initialization, DIAL further improves the macro score by \(5.5\) and \(8.0\) points, respectively. Compared with INTACT~\cite{sun2026intact}, DIAL exceeds its pure CEM and actor CEM variants by \(15.1\) and \(2.2\) points, respectively. The former uses a generic Gaussian proposal, while the latter introduces a separately learned actor for action initialization. DIAL derives its initialization directly from the jointly trained forward and inverse dynamics, without introducing a separate policy training stage.

\begin{figure*}[t]
\centering
\includegraphics[width=\linewidth]{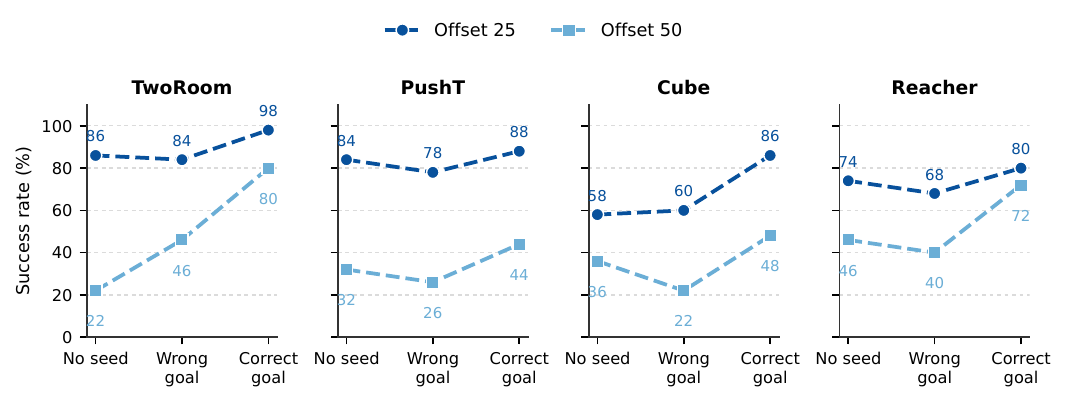}
\vspace{-8mm}
\caption{\textbf{Effect of goal consistency in proposal initialization.} Success rates (\%) for uninitialized, mismatched-goal, and matched-goal proposals across environments and goal offsets under the constrained search budget (\(30\times10\), \(n=50\) episodes per setting).}
\vspace{-4mm}
\label{fig:wrong-goal}
\end{figure*}

\paragraph{Comparison at the 50-step long-horizon offset.} We further evaluate long-horizon planning by doubling the goal offset from \(25\) to \(50\) steps while keeping the evaluation and planning protocols unchanged, as shown in the right part of Table~\ref{tab:combined}. DIAL achieves the highest macro success rate of \(78.0\%\), outperforming Qantara~\cite{rakhimov2026qantara} and LeWM~\cite{maes2026leworldmodel} by \(10.5\) and \(15.0\) points, respectively. The largest gain appears on TwoRoom, where DIAL improves over LeWM by \(36.0\) points. On Cube~\cite{park2025ogbench}, DIAL also exceeds the INTACT actor CEM variant~\cite{sun2026intact} by \(22.0\) points. In contrast, the benefit of improving candidate evaluation alone is limited at this longer horizon. For example, Fast-LeWM-SC~\cite{gao2026fast} improves over Fast-LeWM by only \(1.5\) points. These results indicate that as the temporal distance to the goal increases, providing CEM with a goal-conditioned initial proposal becomes increasingly important for effective search under a fixed planning budget.

\paragraph{Qualitative visualization.}

We provide a qualitative comparison of LeWM~\cite{maes2026leworldmodel}, INTACT~\cite{sun2026intact}, and DIAL in Fig.~\ref{fig:qualitative}. The figure shows representative long-horizon rollouts on Cube~\cite{park2025ogbench} (left) and PushT~\cite{zhou2024dino} (right) under the 50-step goal-offset setting. For each task, the first column shows the initial observation, the intermediate columns show the predicted rollout, and the final column shows the goal observation. LeWM and INTACT either lose effective object interaction or make insufficient progress toward the goal, resulting in rollouts that terminate far from the target. In contrast, DIAL maintains object interaction and makes greater progress toward the goal, consistent with its improved long-horizon performance in Table~\ref{tab:combined}.

\begin{figure*}[t]
\centering
\includegraphics[width=0.99\linewidth]{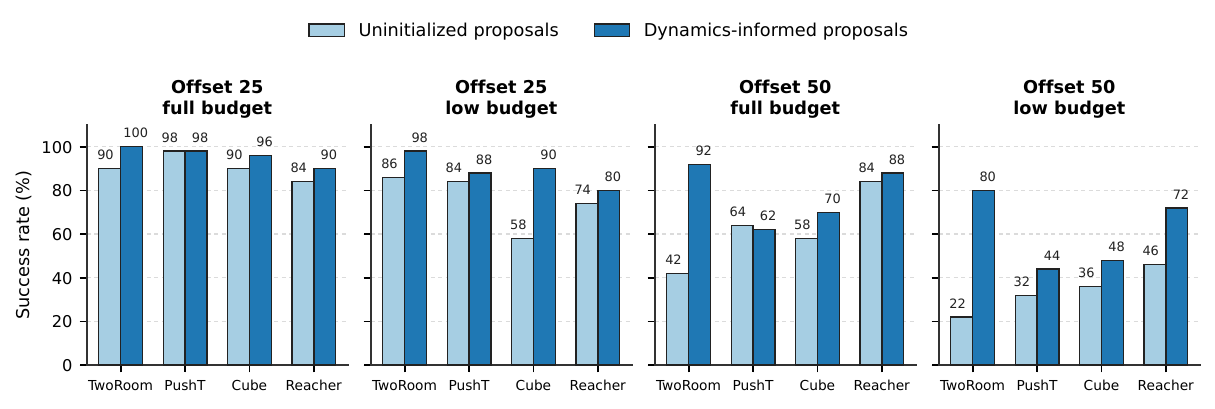}
\vspace{-6mm}
\caption{\textbf{Dynamics-informed initialization under constrained budgets.} Success rates (\%) across environments for uninitialized and dynamics-informed proposals under different goal offsets and search budgets. Full and low budgets correspond to \(300\times30\) and \(30\times10\) candidate rollouts.}
\vspace{-4mm}
\label{fig:seeding-effect}
\end{figure*}

\begin{table*}[t]
\caption{\textbf{Loss ablation at the 25-step goal offset under unseeded CEM.} Each variant is trained from scratch with identical architecture and optimization schedule, and evaluated with the full \(300\times30\) budget. Success rates (\%) are reported for different combinations of training objectives.}
\vspace{-2mm}
\label{tab:ladder}
\centering
\small
\setlength{\tabcolsep}{7pt}
\begin{tabular}{ccccccccc}
\toprule
\(\mathcal{L}_{\mathrm{WM}}\) & \(\mathcal{L}_{\mathrm{RA}}\) & \(\mathcal{L}_{\mathrm{obs}}\) & \(\mathcal{L}_{\mathrm{roll}}\) & PushT & Cube & Reacher & TwoRoom & Macro \\
\midrule
\checkmark &  &  &  & \textbf{96.0} & 72.0 & 80.0 & 84.0 & 83.0 \\
\checkmark & \checkmark &  &  & 94.0 & 70.0 & 88.0 & 82.0 & 83.5 \\
\checkmark &  & \checkmark &  & 90.0 & 80.0 & 90.0 & 80.0 & 85.0 \\
\checkmark & \checkmark & \checkmark &  & 94.0 & \textbf{84.0} & 82.0 & \textbf{86.0} & 86.5 \\
\checkmark & \checkmark & \checkmark & \checkmark & \textbf{96.0} & \textbf{84.0} & \textbf{92.0} & \textbf{86.0} & \textbf{89.5} \\
\bottomrule
\end{tabular}
\vspace{-4mm}
\end{table*}

\subsection{Ablation Study}
\label{sec:ablation}

\paragraph{Effect of goal consistency in proposal initialization.}
To determine whether the benefit of DIAL comes from goal-conditioned initialization rather than simply shifting the proposal distribution away from the default prior, we replace the target latent with one from a different episode while keeping the inverse dynamics module and CEM procedure unchanged. We evaluate this control under the constrained \(30\times10\) search budget. As shown in Fig.~\ref{fig:wrong-goal}, correctly matched goal proposals consistently outperform mismatched proposals across all eight environment and offset settings, with an average gap of \(21.5\) percentage points. In most settings, mismatched proposals provide little or no improvement over uninitialized CEM and can even degrade performance. These results indicate that the benefit of proposal initialization depends on its consistency with the target goal, rather than simply on shifting the proposal mean away from the default prior. The target latent therefore provides essential goal-specific information for constructing an informative initial proposal for CEM.

\paragraph{Dynamics-informed initialization under constrained budgets.}

We evaluate dynamics-informed proposal initialization across search budgets and goal offsets (Fig.~\ref{fig:seeding-effect}). At the 25-step offset, initialization improves mean success by \(13.5\) percentage points under the low budget (\(30\times10\)), compared with \(3.5\) points under the full budget (\(300\times30\)), indicating that its benefit is more pronounced when CEM has fewer opportunities to recover through exploration. The effect grows substantially at the 50-step offset with the planning horizon fixed at \(H=5\), where the mean gains increase to \(27.0\) and \(16.0\) points under the low and full budgets, respectively. In particular, TwoRoom improves from \(22\%\) to \(80\%\) under the low budget. These results indicate that dynamics-informed initialization becomes increasingly valuable when the search budget is constrained or the goal is temporally distant.

\paragraph{Loss ablation for model training.}
To assess the contribution of each training objective, Table~\ref{tab:ladder} reports a controlled ablation at the 25-step goal offset. Starting from the teacher-forced baseline (\(\mathcal{L}_{\mathrm{WM}}\)), adding rollout alignment (\(\mathcal{L}_{\mathrm{RA}}\)) provides only a marginal improvement in macro success, from \(83.0\%\) to \(83.5\%\). In contrast, multi-step inverse supervision (\(\mathcal{L}_{\mathrm{obs}}\)) increases macro success to \(85.0\%\), including an \(8.0\)-point gain on Cube. Combining the two objectives further raises macro success to \(86.5\%\). However, \(\mathcal{L}_{\mathrm{obs}}\) enforces action recoverability only on ground-truth transitions and does not directly constrain model-generated rollouts. \(\mathcal{L}_{\mathrm{roll}}\) addresses this limitation by applying the learned inverse head to autoregressively predicted latent pairs, encouraging model-generated transitions to preserve action-relevant information. Adding \(\mathcal{L}_{\mathrm{roll}}\) yields the highest macro success of \(89.5\%\), with the best or tied-best performance on all four tasks: \(96.0\%\) on PushT, \(84.0\%\) on Cube, \(92.0\%\) on Reacher, and \(86.0\%\) on TwoRoom. The results show that preserving action recoverability on both ground-truth transitions and model-generated rollouts is important for effective planning.

\section{Conclusion}

We present Dynamics-Informed Action Learning (DIAL), a framework that connects latent dynamics learning with goal-conditioned planning through rollout-aligned training and learned inverse dynamics. DIAL trains the latent dynamics model on autoregressive rollouts while learning an inverse dynamics head to preserve action-relevant information in both observed and model-generated transitions. At planning time, the learned inverse and forward dynamics are coupled to construct a dynamics-informed action sequence that initializes the mean of the CEM proposal, while the subsequent search procedure remains unchanged. Across four visual planning benchmarks, DIAL improves goal-reaching performance at both standard \(25\)-step and extended \(50\)-step goal offsets, with particularly large gains under constrained search budgets. These results demonstrate that action-conditioned information learned during latent dynamics training can directly guide goal-conditioned search, improving planning without altering the CEM.
{
    \small
    \bibliographystyle{ieeenat_fullname}
    \bibliography{main}
}

\clearpage
\maketitlesupplementary

The appendix is organized as follows. Sec.~\ref{app:implementation} describes
the model architecture, data preprocessing, optimization, loss configuration,
and other implementation details for reproducing DIAL. Sec.~\ref{app:comparison}
specifies the controlled evaluation protocols and search budgets. Sec.~\ref{app:additional}
provides additional qualitative comparisons, inverse-head diagnostics,
rollout-depth analysis, and protocol sensitivity.

\section{Implementation Details}
\label{app:implementation}

\paragraph{Architecture.}

DIAL is built on an action-conditioned LeWM backbone without a separate
policy network. Each $224\times224$ image is encoded by a ViT-Tiny encoder
with patch size $14$, and its CLS token is projected to a
$D=192$-dimensional latent state. The action encoder maps each normalized
action block to the same dimension using a $1\times1$ temporal convolution
followed by a two-layer SiLU MLP with hidden width $4D=768$. The latent
predictor uses a causal context of at most $h=3$ latent-action pairs. Its
passive branch contains six transformer layers with 16 heads, head dimension
64, MLP width 2048, and dropout 0.1, while the action-conditioned residual
branch contains three transformer layers with a rank-32 bottleneck.

The inverse dynamics head applies layer normalization to an ordered pair of
latent states, followed by a 256-unit SiLU hidden layer and a linear
projection to the action dimension. It contains 0.102M parameters. The
predictor supports variable-length contexts up to $h=3$ and predicts the next
latent state at every valid position. All visual, action, dynamics, and
inverse modules are optimized jointly.

\paragraph{Data preprocessing.}

Following prior work~\cite{maes2026leworldmodel,gao2026fast}, we use the
original offline datasets and success criteria of each benchmark. RGB
observations are resized to $224\times224$ and normalized using ImageNet
statistics. Actions are normalized using statistics computed from the
corresponding dataset. Five consecutive primitive actions are concatenated
into one action block, matching the frame skip of five. Unless otherwise
specified, all experiments use clean observations without synthetic
distractors.

\paragraph{Optimization and loss configuration.}

Table~\ref{tab:implementation} summarizes the training configuration. Models
are trained for 10 epochs with a
learning rate of $5\times10^{-5}$, weight decay $10^{-3}$, batch size 128,
linear warmup followed by cosine annealing, and gradient clipping with
maximum norm 1.0. We use a $90/10$ train/validation split. The normalized
weights for the observed inverse loss are
$(\omega_1,\omega_2,\omega_3)=(3,2,1)/6$. The predicted-transition inverse
loss is disabled for the first three epochs and then applied with
$k\leq K_{\mathrm{roll}}=2$. Unless otherwise specified, all DIAL-specific
loss coefficients are set to
$\beta_{\mathrm{RA}}=\beta_{\mathrm{obs}}=\beta_{\mathrm{roll}}=1$ across
all environments.

\begin{table*}[h]
\centering
\small
\caption{\textbf{DIAL training hyperparameters.}
Values are shared across PushT, Cube, Reacher, and TwoRoom unless noted otherwise.}
\label{tab:implementation}
\begin{tabular}{lll}
\toprule
component & hyperparameter & value \\
\midrule
input & image resolution / patch size & \(224\times224\) / \(14\) \\
representation & latent dimension \(D\) & \(192\) \\
context & history \(h\) / rollout depth \(L\) & \(3\) / \(5\) \\
predictor & layers / heads / MLP width & \(6\) / \(16\) / \(2048\) \\
residual branch & layers / bottleneck rank & \(3\) / \(32\) \\
inverse head & depth / hidden width & \(2\) / \(256\) \\
\midrule
optimization & epochs / batch size & \(10\) / \(128\) \\
optimization & optimizer / learning rate & AdamW / \(5\times10^{-5}\) \\
optimization & learning-rate schedule & linear warmup and cosine annealing \\
optimization & weight decay / gradient clip & \(10^{-3}\) / \(1.0\) \\
optimization & precision / train split & bfloat16 / \(0.9\) \\
\midrule
SIGReg & weight / knots / projections & \(0.09\) / \(17\) / \(1024\) \\
rollout alignment & \(\beta_{\mathrm{RA}}\) / \(L\) & \(1.0\) / \(5\) \\
observed inverse & \(\beta_{\mathrm{obs}}\) / \(K_{\mathrm{obs}}\) & \(1.0\) / \(3\) \\
predicted inverse & \(\beta_{\mathrm{roll}}\) / \(K_{\mathrm{roll}}\) & \(1.0\) / \(2\) \\
predicted inverse & warmup & first \(3\) epochs \\
\bottomrule
\end{tabular}
\end{table*}

\section{Comparison Protocol}
\label{app:comparison}

The offset-25 results in Table~\ref{tab:implementation} follow the reported planning protocols of the corresponding methods and therefore do not use a uniform search budget. We use these results as the standard-offset reference for comparison with prior work. For the controlled offset-50 evaluation, all search-based methods use the same CEM configuration, with 300 candidates per iteration, 30 iterations, 30 elites, and initial variance 1.0. All methods are evaluated on identical start-goal pairs under the same planning horizon and action block size. The low-budget setting used in the ablation studies reduces the search to 30 candidates per iteration, 10 iterations, and 3 elites. Apart from the search budget, the planning objective and evaluation protocol remain unchanged. This controlled setting isolates the effect of proposal initialization from differences in search budget. All methods use a planning horizon of \(H=5\), with each action block corresponding to five environment steps. The optimized sequence is executed before replanning. Unless otherwise specified, replanning therefore occurs every 25 environment steps.

\section{Additional Experimental Results}
\label{app:additional}

\begin{table*}[!t]
\caption{\textbf{Sensitivity to replanning interval and warm starting.}
Macro success at the 50-step goal offset for standard CEM and
dynamics-informed initialization under different replanning intervals.
Entries are reported as standard CEM $\rightarrow$ dynamics-informed
initialization.}
\label{tab:recede}
\centering
\small
\begin{tabular}{llccccc}
\toprule
\(r\) & warm start & TwoRoom & PushT & Cube & Reacher & macro \\
\midrule
5 & on  & \(42\to92\) & \(62\to58\) & \(54\to58\) & \(88\to84\) & \(61.5\to73.0\) \\
1 & on  & \(44\to86\) & \(30\to28\) & \(50\to58\) & \(38\to46\) & \(40.5\to54.5\) \\
1 & off & \(22\to90\) & \(30\to46\) & \(54\to60\) & \(30\to82\) & \(34.0\to69.5\) \\
\bottomrule
\end{tabular}
\end{table*}

\subsection{Qualitative Comparison}
\label{app:qualitative}

\begin{figure*}[p]
\centering
\begin{minipage}[t][0.48\textheight][t]{0.49\textwidth}
\centering
\includegraphics[width=\linewidth,height=0.36\textheight,keepaspectratio]{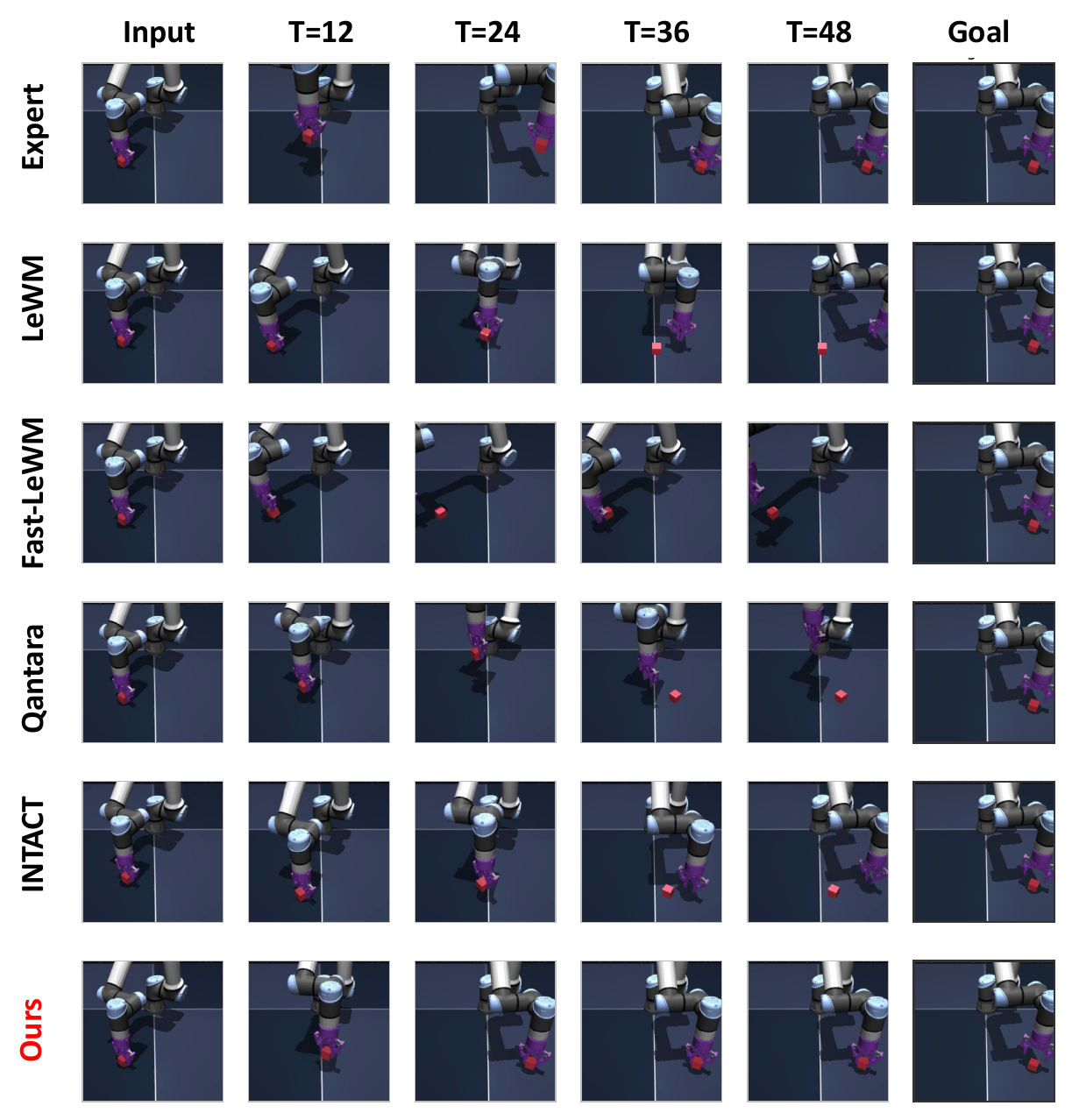}
\caption{\textbf{Qualitative comparison on Cube under the 50-step goal offset.}
The expert row shows the dataset trajectory from the same initial state to the
given goal. LeWM, Fast-LeWM, Qantara, and INTACT (Pure CEM) fail to complete
the required manipulation in this episode, while DIAL maintains contact with
the cube throughout the rollout and reaches the target configuration.}
\label{fig:qual-cube-full}
\end{minipage}\hfill
\begin{minipage}[t][0.48\textheight][t]{0.49\textwidth}
\centering
\includegraphics[width=\linewidth,height=0.36\textheight,keepaspectratio]{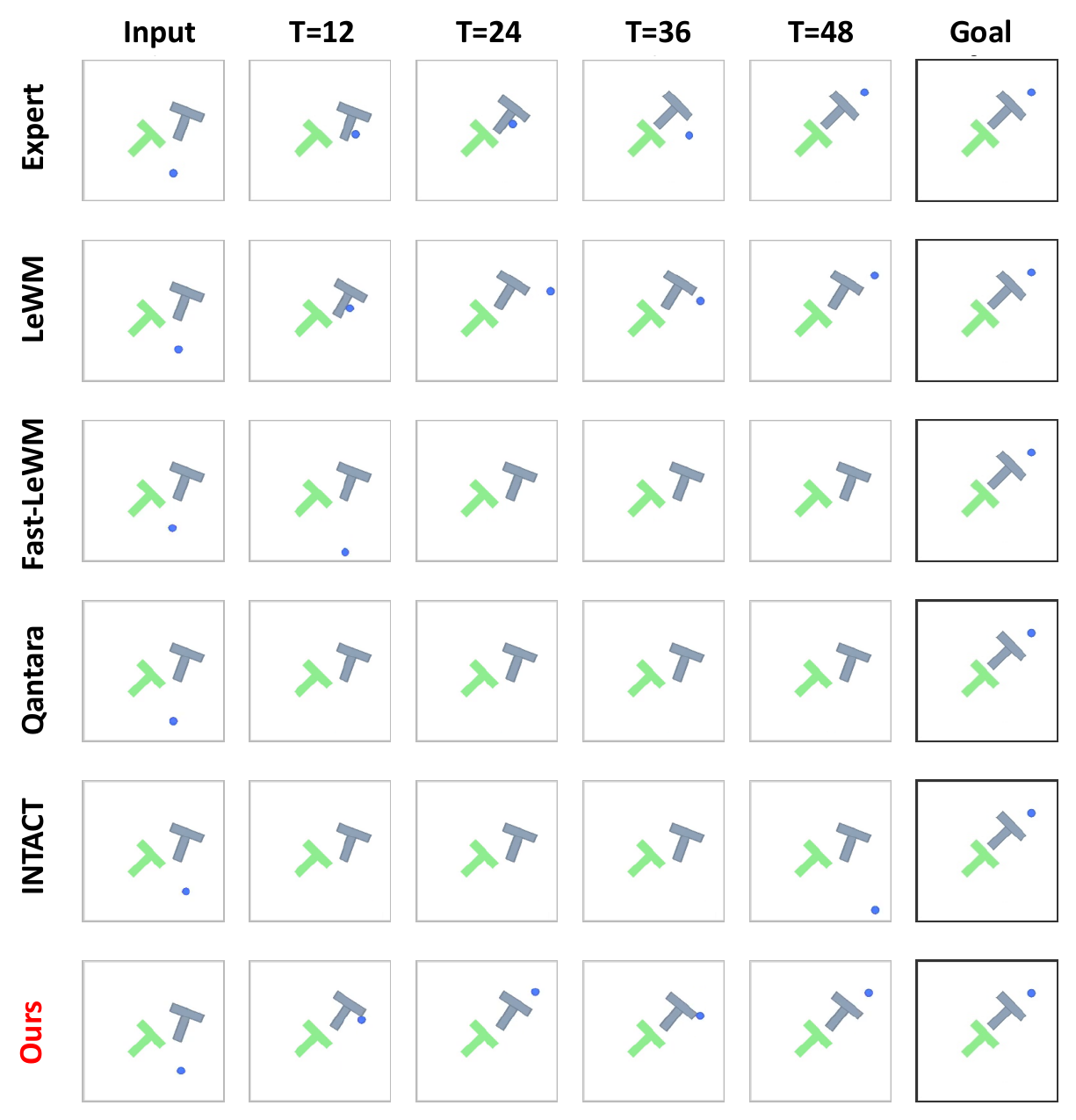}
\caption{\textbf{Qualitative comparison on PushT under the 50-step goal offset.}
The expert row shows the dataset trajectory for the same initial state and
goal. The four external baselines fail to reach the target arrangement over
the long-horizon rollout, whereas DIAL maintains effective object interaction
and progressively moves the object toward the desired configuration.}
\label{fig:qual-pusht-full}
\end{minipage}

\vfill

\begin{minipage}[t][0.48\textheight][t]{0.49\textwidth}
\centering
\includegraphics[width=\linewidth,height=0.36\textheight,keepaspectratio]{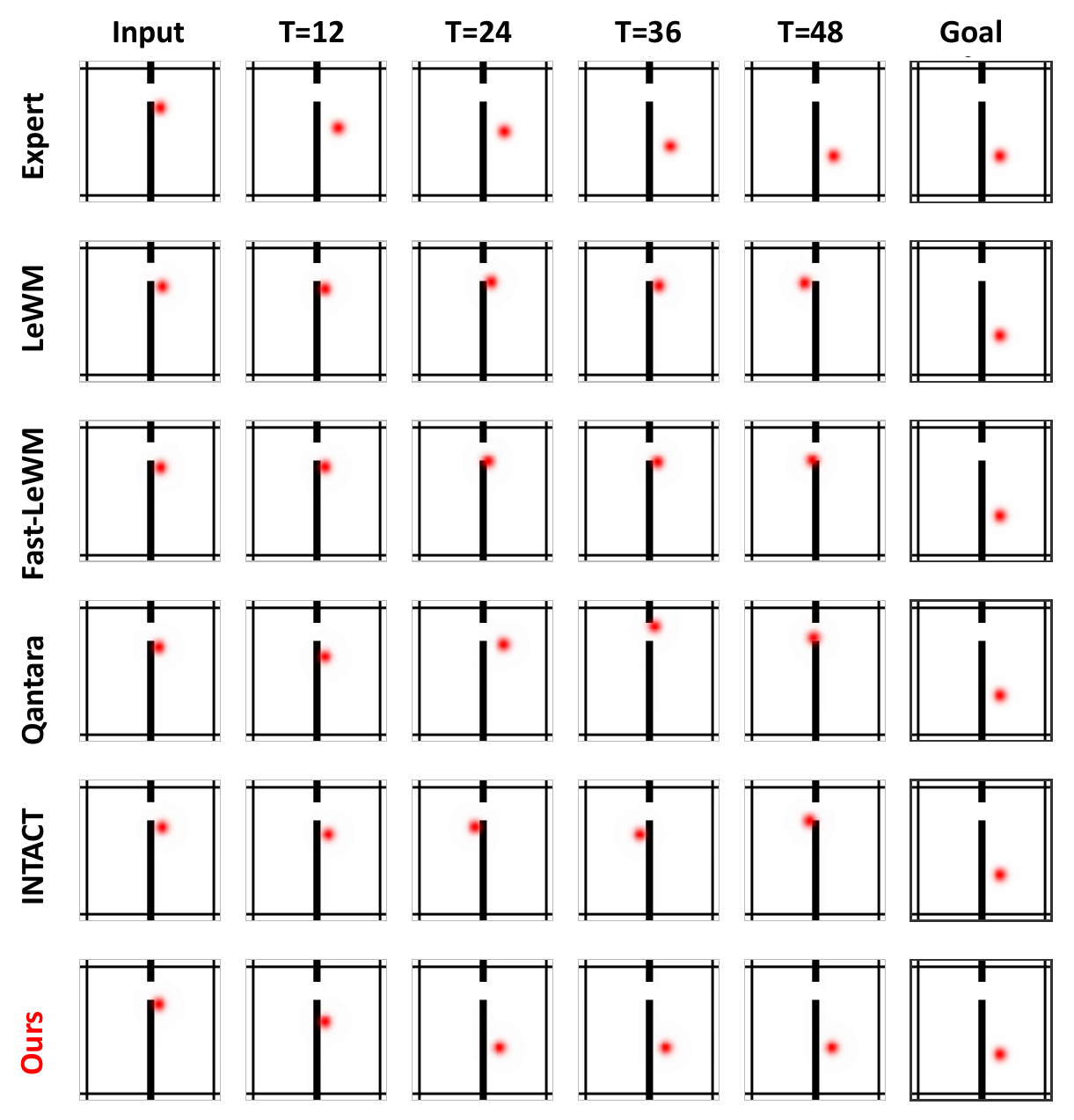}
\caption{\textbf{Qualitative comparison on TwoRoom under the 50-step goal offset.}
The expert row shows the dataset trajectory for the same initial state and
goal. The external baselines remain in the starting room and fail to reach
the goal region, whereas DIAL successfully passes through the doorway and
continues toward the target room.}
\label{fig:qual-tworoom-full}
\end{minipage}\hfill
\begin{minipage}[t][0.48\textheight][t]{0.49\textwidth}
\centering
\includegraphics[width=\linewidth,height=0.36\textheight,keepaspectratio]{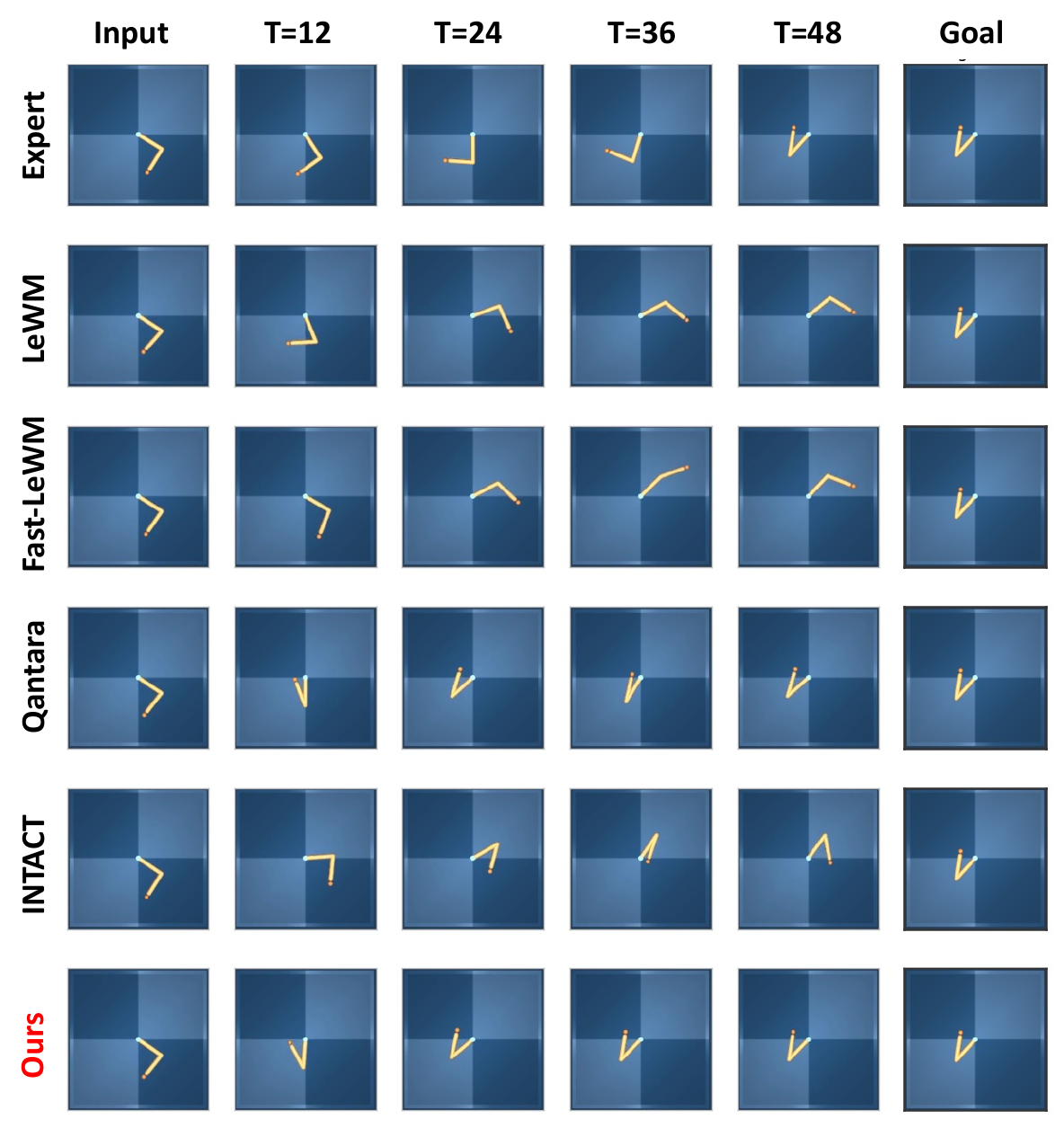}
\caption{\textbf{Qualitative comparison on Reacher under the 50-step goal offset.}
The expert row shows the dataset trajectory for the same initial state and
goal. LeWM, Fast-LeWM, Qantara, and INTACT (Pure CEM) fail the success
criterion on this episode, while DIAL produces a rollout that reaches the
target pose.}
\label{fig:qual-reacher-full}
\end{minipage}
\end{figure*}

We provide additional qualitative comparisons under the 50-step goal-offset
setting, including LeWM~\cite{maes2026leworldmodel}, Fast-LeWM~\cite{gao2026fast},
Qantara~\cite{rakhimov2026qantara}, and INTACT (Pure CEM)~\cite{sun2026intact}.
As shown in Figs.~\ref{fig:qual-cube-full} to
\ref{fig:qual-reacher-full}, the first column shows the initial observation,
the intermediate columns show the predicted rollout, and the final column
shows the goal observation. Across the four environments, DIAL produces
rollouts that remain more goal-directed over the long horizon, consistent
with its improved performance under the 50-step offset.

\begin{table}[h]
\caption{\textbf{Inverse-head fidelity at different temporal gaps.}
Normalized $\ell_2$ error, $R^2$, and maximum predicted action magnitude of
$G_\psi$ at five- and ten-step gaps across four environments.}
\label{tab:mismatch}
\begin{center}
\small
\resizebox{\linewidth}{!}{%
\begin{tabular}{lcccccc}
\toprule
& \multicolumn{3}{c}{\(k{=}5\)} & \multicolumn{3}{c}{\(k{=}10\)} \\
\cmidrule(lr){2-4}\cmidrule(lr){5-7}
env & nl2 & \(R^2\) & \(\|\hat a\|_\infty\) & nl2 & \(R^2\) & \(\|\hat a\|_\infty\) \\
\midrule
TwoRoom & 0.98 & \(-0.02\) & 1.3 & 1.07 & \(-0.24\) & 2.3 \\
Reacher & 1.09 & \(-0.21\) & 0.8 & 1.14 & \(-0.32\) & 0.8 \\
Cube    & 1.94 & \(-2.13\) & 4.3 & 2.36 & \(-4.55\) & 5.2 \\
PushT   & 4.61 & \(-13.4\) & 4.1 & 5.62 & \(-16.8\) & 4.5 \\
\bottomrule
\end{tabular}%
}
\end{center}
\end{table}

\subsection{Effect of Rollout Depth}
\label{app:rollout-depth}

We further examine the effect of the rollout-alignment depth $L$. Increasing
$L$ from $5$ to $10$ does not improve planning performance. On PushT at the
50-step goal offset, success decreases from $58\%$ to $34\%$, approaching the
performance of the model trained without rollout alignment. We therefore use
$L=5$ throughout the main experiments, matching the planning horizon.

\subsection{Inverse-head Fidelity}
\label{app:ablation}

Table~\ref{tab:mismatch} evaluates the inverse head $G_\psi$ at the temporal
gaps used to construct the dynamics-informed action sequence. At a five-step
gap, the normalized $\ell_2$ error ranges from $0.98$ on TwoRoom to $4.61$ on
PushT, while $R^2$ is negative across all four environments. Increasing the
gap to ten steps further degrades inverse prediction, with particularly low
$R^2$ on Cube ($-4.55$) and PushT ($-16.8$). The predicted action magnitude
also exceeds the normalized action bound in TwoRoom, Cube, and PushT, while
remaining within the bound on Reacher. These results indicate that the
inverse head is not a uniformly accurate action predictor, particularly at
larger temporal gaps. Instead, its predictions provide a coarse
goal-directed initialization that is subsequently refined by CEM.

\subsection{Protocol Sensitivity}
\label{app:protocol}

Table~\ref{tab:recede} examines DIAL under different replanning intervals at
the 50-step goal offset. With the default $r=5$ setting and warm starting,
macro success increases from $61.5\%$ with standard CEM to $73.0\%$ with
dynamics-informed initialization. Replanning after every action block
($r=1$) reduces macro success to $40.5\%$ with standard CEM and $54.5\%$ with
dynamics-informed initialization. When warm starting is disabled at $r=1$,
macro success further decreases to $34.0\%$ with standard CEM, while
dynamics-informed initialization achieves $69.5\%$. These results show that
the benefit of dynamics-informed initialization remains substantial even
without carrying the previous solution into the next planning step. They also
indicate that more frequent replanning does not necessarily improve
performance under a fixed planning budget, supporting the default $r=5$
protocol used in our experiments.

\end{document}